\documentclass[letterpaper, 10 pt, conference]{ieeeconf}  

\IEEEoverridecommandlockouts                              
 
\usepackage{graphics} 
\usepackage{epsfig} 
\usepackage{mathptmx} 
\usepackage{times} 
\usepackage{amsmath} 
\usepackage{amssymb}  
\usepackage{arydshln}
\usepackage{booktabs}
\usepackage{pifont}
\usepackage{tabularx}
\usepackage[table]{xcolor}
\usepackage[dvipsnames]{xcolor}
\usepackage{verbatim}
\usepackage{multirow}
\usepackage{makecell}
\usepackage{cite}
\usepackage{pifont}
\usepackage{float}  
\usepackage{balance}
\usepackage{caption}

\newcommand{\methodname}{DexPIE}

\definecolor{rblue}{rgb}{0,0.5,1}
\definecolor{awesome}{rgb}{1.0, 0.13, 0.32}
\definecolor{hollywoodcerise}{rgb}{0.96, 0.0, 0.63}
\definecolor{lasallegreen}{rgb}{0.03, 0.47, 0.19}
\definecolor{hanpurple}{rgb}{0.32, 0.09, 0.98}
\definecolor{green(pigment)}{rgb}{0.0, 0.65, 0.31}

\makeatletter
\let\NAT@parse\undefined
\makeatother
\usepackage[pagebackref=false, breaklinks=true, colorlinks, bookmarks=false]{
        hyperref
}
\hypersetup{
        colorlinks=true,
        linkcolor={red},
        citecolor={hanpurple},
        urlcolor={magenta}
}

\title{\LARGE \bf
DexPIE: Stable Dexterous Policy Improvement\\from 
Real-World Experience
}

\author{Ruizhe Liao$^{1}$ \quad Wenrui Chen$^{1, \dagger}$ \quad Liangji Zeng$^{1}$ \quad Haoran Lin$^1$ \quad Fan Yang$^1$\\Kailun Yang$^1$ \quad Yaonan Wang$^{1}$ \\[4pt]
$^1$Hunan University \quad $^\dagger$Corresponding author\\[2pt]
{\hypersetup{urlcolor=blue}\url{https://siiuuuuuu.github.io/DexPIE}}}

\begin{document}

\maketitle
\thispagestyle{empty}
\pagestyle{empty}

\begin{abstract}
Dexterous manipulation presents substantial challenges for imitation learning due to its high-dimensional action space and complex contact-rich dynamics. Policies trained purely from demonstrations often suffer from compounding errors during deployment and require large amounts of expert data to achieve reliable performance. To move beyond the limitations of demonstration data, in this work, we propose \methodname{}, a post-training framework for dexterous policy improvement from experience collected through real-world deployment. First, \methodname{} enables effective exploration coverage through a dexterous-hand-adapted intervention system and multi-stage DAgger-style data collection across initial and intermediate task stages. Meanwhile, we enhance consistency between training and inference to reduce the distribution shift between rollouts and demonstration data, better aligning rollout behavior with demonstrations, allowing the critic to learn a value function induced by a more consistent underlying policy. Together, these components provide reliable supervision for policy evaluation. Finally, \methodname{} improves the policy through conditioning on a continuous optimality indicator, allowing the policy to leverage the quality of data in a more fine-grained manner. Across three challenging real-world dexterous manipulation tasks, \methodname{} achieves a 37.3\% improvement in success rate over the demonstration-based reference policy, outperforming all baseline methods and demonstrating stronger robustness. The source code and dataset will be made publicly available.
\end{abstract}
\section{Introduction}

\begin{figure}[!t]
    \centering
    \includegraphics[width=\linewidth]{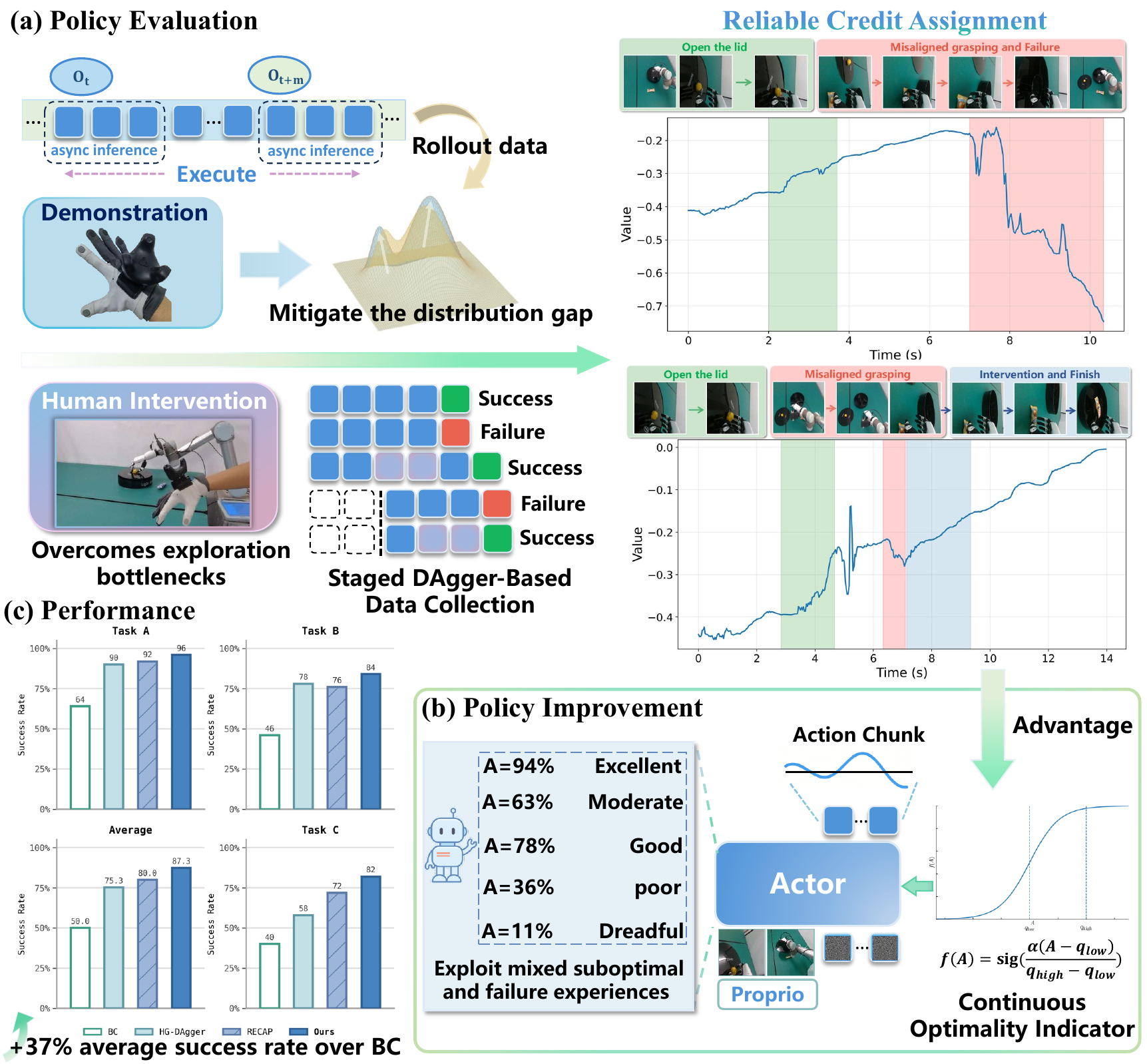}
    \caption{\textbf{DexPIE overview}. We improve both policy evaluation and policy improvement to reliably exploit mixed suboptimal and failure experience in real-world dexterous manipulation. (a) \textbf{Policy Evaluation}: We first improve training–inference consistency to mitigate the distribution gap between rollout and demonstration data, and then perform staged DAgger-style data collection with a dexterous-hand-adapted human intervention system. These two components enable reliable credit assignment that finely distinguishes progressive, stagnant, and failure-inducing experiences. (b) \textbf{Policy Improvement}: We introduce a continuous optimality function that enables stable optimality-conditioned policy improvement by preserving the relative ordering of action quality instead of collapsing it into binary labels. 
    (c) \textbf{Performance}: \methodname{} achieves over 30\% success-rate gains over the demonstration-based policy across all three tasks.}
    \label{fig:teaser}
\end{figure}

Imitation Learning (IL) has made significant progress in robotic manipulation by learning from demonstrations~\cite{zhao2023learning,chi2025diffusion}.
Beyond the requirement for large-scale demonstration data, IL suffers from compounding errors during deployment~\cite{osa2018algorithmic}, which often leads to performance saturation.
Moreover, simply incorporating DAgger-style post-training remains within the imitation-learning paradigm.
Reinforcement Learning (RL)~\cite{sutton1998reinforcement}, in contrast, provides a promising alternative for enabling policies to improve autonomously through real-world deployment beyond expert demonstrations.
This suggests that effective robot learning should not only leverage demonstration data, but also exploit experience collected from real-world rollouts, allowing the policy to correct deployment-time errors, complete tasks more robustly, and ultimately achieve self-improvement~\cite{intelligence2025pi06,chen2025conrft,li2025gr,yang2026aloe}.

Extending this paradigm to dexterous manipulation, however, remains challenging.
Recent works~\cite{chen2025conrft,intelligence2025pi06,luo2025precise,yang2026aloe,xu2024rldg} have shown that Human-In-the-Loop (HIL) RL post-training can effectively address exploration bottlenecks of policies trained from demonstrations, substantially alleviate Out-Of-Distribution (OOD) issues, and improve sample efficiency.
Nevertheless, most of these methods are limited to parallel grippers and cannot be directly adapted to dexterous hands.
Furthermore, effective RL for long-horizon manipulation requires informative and fine-grained reward signals.
Sparse rewards, such as binary success indicators, are widely used in real-world reinforcement learning due to their simplicity and generality, but they introduce severe credit-assignment challenges in long-horizon tasks. 
Existing works~\cite{chen2025sarm,zhang2025rewind,mao2026arm} mainly address this issue by manually or automatically annotating task stages and progress labels, \textit{e.g.}, with human or LLM supervision.
Recent work~\cite{intelligence2025pi06,yu2026kai0} achieves more stable and scalable policy improvement through advantage-conditioned policy extraction.
Nevertheless, they use binary optimality labels, which fail to preserve the relative ordering of action quality.
In addition, previous research works~\cite{yu2026kai0,black2025training,tang2025vlash,black2026RTC} show that improving train-inference consistency through asynchronous inference can enhance policy performance. 
We further observe that the resulting inconsistent deployment behaviors introduce distribution shift between demonstrations and deployment rollouts. 
We refer to this mismatch as the demonstration-deployment gap. 
Such a gap not only provides low-quality data for imitation learning but also leads to a heterogeneous mixture of mismatched behaviors in the post-training dataset, making critic learning difficult.

As shown in Fig.~\ref{fig:teaser}, to address these challenges, we propose \methodname{}. For reliable policy evaluation, we improve training--inference consistency to mitigate the distribution gap between rollout and demonstration data, and combine staged DAgger-style data collection with a dexterous-hand-adapted human intervention system.
Together, these components enable reliable credit assignment.
For policy improvement, we introduce continuous optimality conditioning that preserves the relative ordering of action quality instead of collapsing diverse behaviors into binary labels.
Across three real-world dexterous manipulation tasks, \methodname{} achieves over 30\% success-rate gains over the demonstration-based policy.

\section{Related Work}

\textbf{Human-in-the-Loop Robot Learning.} 
Imitation learning collects expert demonstrations through leader--follower robotic arms, VR devices, exoskeletons, and other teleoperation interfaces~\cite{zhao2023learning,wu2024gello,yang2024ace,wang2024dexcap,qin2023anyteleop}.
However, policies trained purely from expert demonstrations often fail to recover from errors accumulated during online execution.
To address this issue, classical approaches build upon DAgger~\cite{ross2011reduction}.
HG-DAgger~\cite{kelly2019hg} further allows a teleoperator to intervene when the policy enters undesirable states.
Human-in-the-loop mechanisms have also demonstrated their effectiveness in real-world RL.
Under sparse reward settings, human intervention can accelerate early-stage policy exploration and alleviate the exploration challenges faced by demonstration-based policies~\cite{luo2025precise,chen2025conrft,xu2024rldg,intelligence2025pi06}.
Recent systems~\cite{hu2025rac,wu2025robocopilot,evorl2026} improve teleoperation and intervention interfaces to enable intuitive intervention for gripper--arm systems.
For dexterous manipulation systems, DexGrasp-VLA~\cite{cui2025end} and DexHiL~\cite{han2026dexhil} enable intervention through shared autonomy assisted by an autonomous hand policy and hand retargeting, respectively.

\textbf{Algorithms and Systems for Real-World RL.} 
Recent works have achieved promising policy improvement in simulation environments using PPO and its variants~\cite{li2025simplevla,zhai2025VLAC,chen2025pirl}.
However, due to the low sample efficiency and high noise of real-world interaction, such methods remain difficult to transfer to real-world scenarios.
Meanwhile, some works focus on reward specification or progress modeling in the real world through manually designed reward functions, VLM-based reward annotation, or stage-wise labeling~\cite{ma2025VLM,zhai2025VLAC,zhang2025rewind,mao2026arm,chen2025sarm}. 
Since real-world datasets are often composed of multi-source heterogeneous data, another line of work studies offline and off-policy reinforcement learning in real-world settings~\cite{yang2026aloe,chen2025conrft,xu2024rldg,luo2025precise,li2025gr,huang2025co}, which can efficiently reuse historical trajectory data and improve sample efficiency, leading to significant progress in real-world RL. Auxiliary-module methods~\cite{wagenmaker2025DSRL,xiao2026PLD} adapt a pretrained policy with lightweight components.  However, improvement is limited by the auxiliary module.
RECAP~\cite{intelligence2025pi06} improves the policy through binary advantage-label conditioning, achieving significant performance gains on real-world long-horizon tasks. 
Compared with binary labels, our continuous optimality function captures action quality more finely.

\section{Preliminaries}
\textbf{Reinforcement Learning.}
We consider the standard Reinforcement Learning (RL) setting~\cite{levine2018reinforcement}, where an agent, represented by a policy $\pi(a_t \mid o_t)$, selects an action $a_t \in \mathcal{A}$ given an observation $o_t \in \mathcal{O}$. A trajectory is denoted by $\tau=(o_0,a_0,\ldots,o_T)$, and the induced trajectory distribution is $\rho_{\pi}(\tau)=p(o_0)\prod_{t=0}^{T-1}\pi(a_t\mid o_t)p(o_{t+1}\mid o_t,a_t)$. The reward function is given by $r(o_t,a_t)$, abbreviated as $r_t$, and $\gamma\in[0,1]$ denotes the discount factor. 
The discounted cumulative reward, or return, is defined as $R(\tau)=\sum_{t=0}^{T}\gamma^t r_t$. The goal of RL is to learn a policy that maximizes the expected return, \textit{i.e.}, $J(\pi)=\mathbb{E}_{\tau\sim\rho_{\pi}(\tau)}[R(\tau)]=\mathbb{E}_{\tau\sim\rho_{\pi}(\tau)}[\sum_{t=0}^{T}\gamma^t r_t]$. The value function of policy $\pi$ is then defined as $V^{\pi}(o_t)=\mathbb{E}_{\tau_{t:T}}\left[\sum_{l=t}^{T}\gamma^{l-t}r_l\right]$. In this work, we estimate the advantage of action $a_t$ at observation $o_t$ using an $N$-step return, written as $A^{\pi}(o_t,a_t)=\mathbb{E}_{\rho_{\pi}(\tau)}\left[\sum_{l=t}^{t+N-1}\gamma^{l-t}r_l+\gamma^N V^{\pi}(o_{t+N})\right]-V^{\pi}(o_t)$.

\textbf{Product-Policy View of Optimality Guidance.}
\label{sec:Preliminaries}
To achieve stable improvement over a reference policy $\pi_{\mathrm{ref}}$, we adopt the product-policy view, which performs policy improvement by 
constructing an improved target distribution around the reference policy. 
Specifically, rather than directly optimizing a regularized RL objective, 
we define the target policy $\hat{\pi}$ by reweighting $\pi_{\mathrm{ref}}$ 
with an optimality indicator $I_t$:
\begin{equation}
\hat{\pi}(a_t \mid o_t)
\propto
\pi_{\mathrm{ref}}(a_t \mid o_t)
\cdot
p\!\left(I \mid o_t, a_t\right)^{\beta},
\label{eq:target_distribution}
\end{equation}
where $\beta$ denotes the guidance strength, and 
$p\!\left(I \mid o_t,a_t\right)
=
\frac{
f(A^{\pi_{\mathrm{ref}}}(o_t,a_t))
}{
Z(o_t)
}$, where $Z(o_t)>0$ is an action-independent normalizing constant. Following the product-policy improvement theorem in CFGRL~\cite{frans2025diffusion}, 
improvement over $\pi_{\mathrm{ref}}$ is guaranteed when $f$ is chosen as a non-negative, monotonically increasing function of 
$A^{\pi_{\mathrm{ref}}}(o_t,a_t)$.
Taking the logarithm of Eq.~\eqref{eq:target_distribution} and differentiating with respect to the action yields an additive decomposition of the product-policy score:
\begin{equation}
\nabla_{a_t}\log \hat{\pi}(a_t \mid o_t)
=
\nabla_{a_t}\log \pi_{\mathrm{ref}}(a_t \mid o_t)
+
\beta \nabla_{a_t}\log p(I \mid o_t,a_t).
\label{eq:product_score}
\end{equation}
By applying Bayes' rule to the optimality indicator 
$p(I \mid o_t,a_t)$, its action-gradient can be written as the difference between the optimality-conditioned score and the unconditional score. Substituting this result into Eq.~\eqref{eq:product_score}, we obtain the guided score:
\begin{equation}
\begin{aligned}
\nabla_{a_t}\log \hat{\pi}(a_t \mid o_t)
={}&
(1-\beta)\nabla_{a_t}\log \pi_{\mathrm{ref}}(a_t \mid o_t)
\\
&+
\beta \nabla_{a_t}\log \pi_{\mathrm{ref}}(a_t \mid o_t,I).
\end{aligned}
\label{eq:guided_score}
\end{equation}
This is analogous to Classifier-Free Guidance (CFG)~\cite{ho2022classifier}.
In practice, we implement this guidance in a diffusion policy by conditioning
the denoising network on the optimality indicator $I$.
\label{sec}
\begin{figure*}
    \centering
    \includegraphics[width=0.98\textwidth]{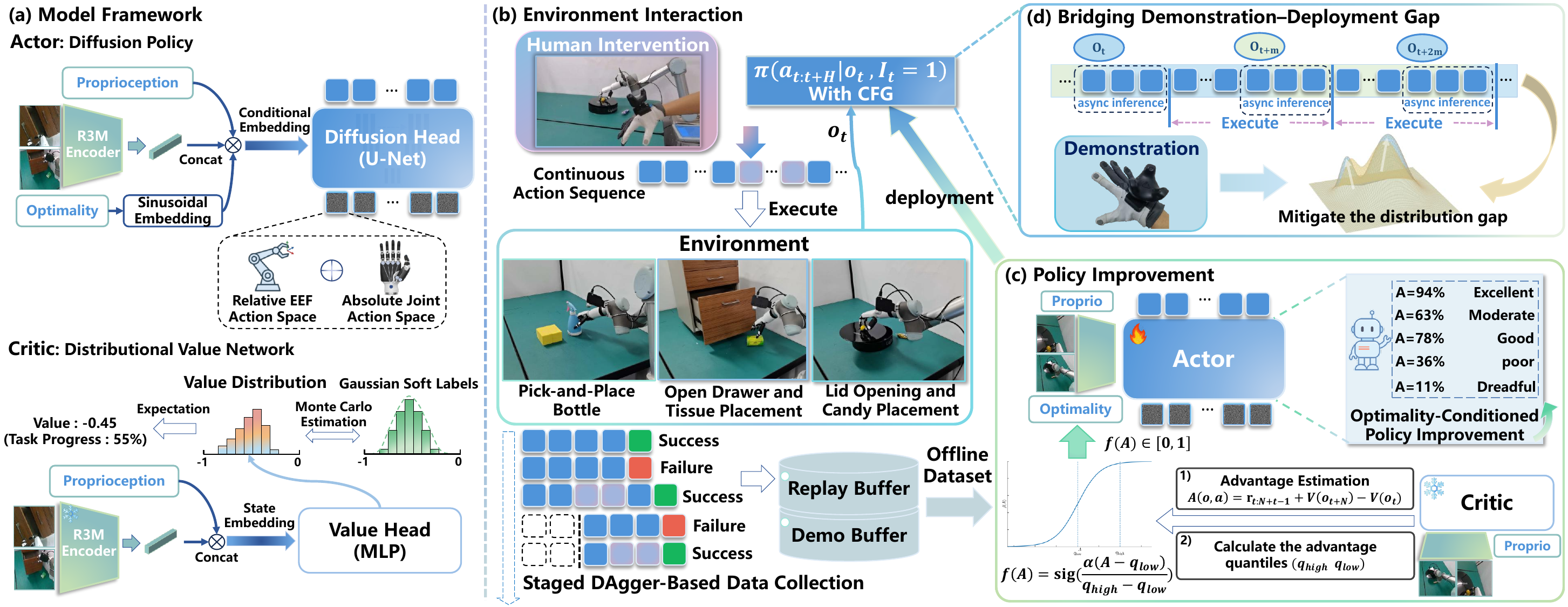}
    \vskip-1ex
    \caption{\textbf{Overview of \methodname{} framework.} 
    \textbf{(a)} The model architecture consists of an actor and a critic. 
    The actor is an optimality-conditioned diffusion policy with a U-Net architecture~\cite{ze2025generalizable}, with an action space defined as relative EEF actions~\cite{chi2024universal} concatenated with absolute dexterous-hand joint actions, while the critic is a distributional value network. 
    \textbf{(b)} The policy is warm-started with demonstration data, and CFG is applied to the optimality indicator. During deployment, failures or exploration bottlenecks trigger human intervention. We further perform DAgger-style data collection across different task stages. \textbf{(c)} After obtaining stable advantage estimates, we perform policy improvement conditioned on a continuous optimality indicator. \textbf{(d)} We reduce the distribution shift between post-training data and demonstrations through improving training–inference consistency.
    }
    \label{fig:framework}
    \vskip-3ex
\end{figure*}

\section{Method}
As shown in Fig.~\ref{fig:framework}, our method aims to provide a complete post-training pipeline for dexterous manipulation. 
In Sec.~\ref{sec:data collection}, we integrate a human-in-the-loop data collection pipeline for dexterous manipulation. 
In Sec.~\ref{sec:async inference}, we improve training–inference consistency to mitigate the demonstration-deployment gap. 
In Sec.~\ref{sec:policy improvement}, based on the offline dataset, we first train the critic using Monte Carlo return. 
After obtaining advantage estimates, we perform policy improvement conditioned on the continuous optimality indicator.

\subsection{Human-in-the-Loop Data Collection for Dexterous Manipulation}
\label{sec:data collection}

\begin{figure}[!t]
    \centering
    \includegraphics[width=0.90\linewidth]{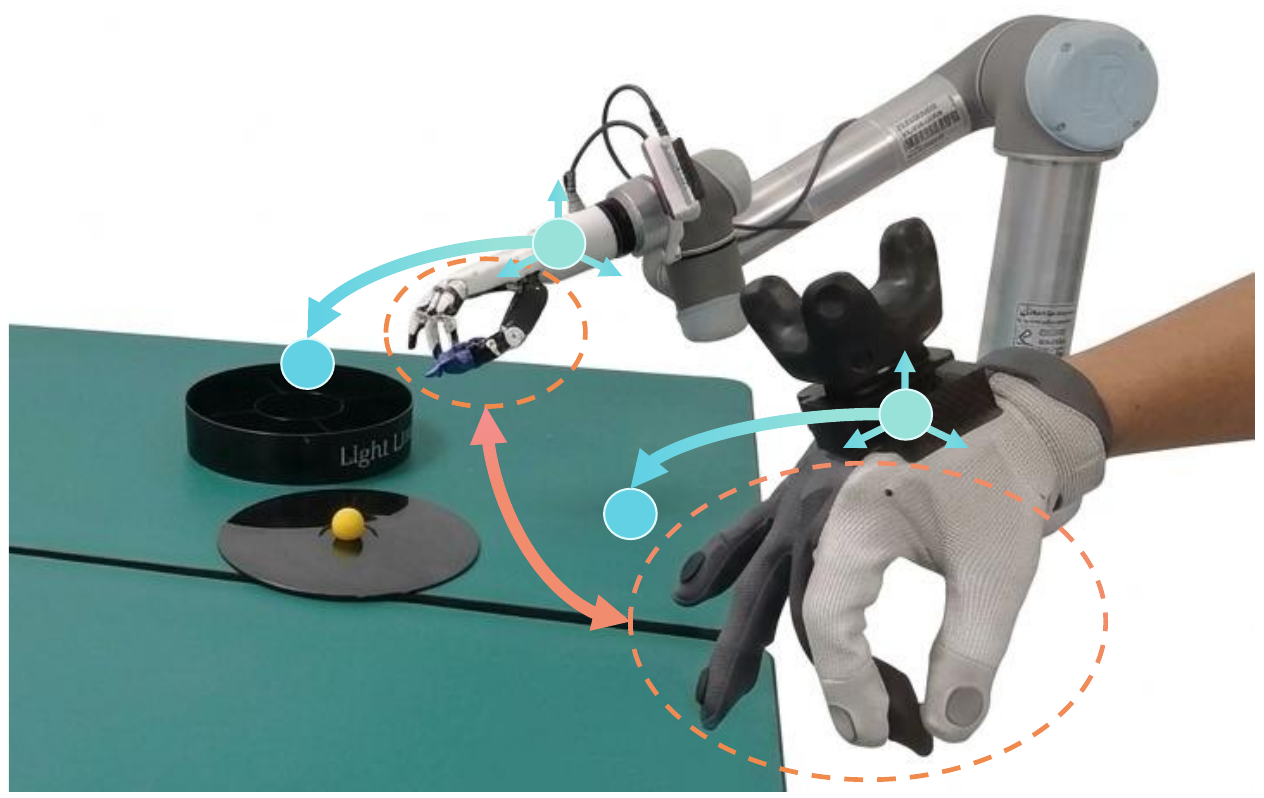}
    \vspace{-1ex}
    \caption{\textbf{Human-as-follower strategy.}
    The operator aligns the wrist orientation and hand gesture with the robot end-effector and dexterous-hand posture before takeover, thereby enabling intuitive human corrective control to be initiated from arbitrary robot states.}
    \label{fig:intervention}
    \vspace{-3ex}
\end{figure}

\textbf{Human-Following Intervention System.}
To provide simple and intuitive corrections during policy deployment for error recovery and exploration bottlenecks, we implement a system integration that enables seamless switching between arm-hand teleoperation and policy execution. 
The system adopts a multi-threaded architecture, where the teleoperation system and the autonomous policy run in parallel, and execution is switched via keyboard commands. 
For the teleoperation interface, we use a Vive tracker to provide the wrist pose and Manus gloves to capture the full hand posture, which is then retargeted to the absolute joint space of the Inspire hand. 
To enable intuitive corrective interventions from arbitrary robot states, we introduce a simple human-as-follower strategy, inspired by the intervention mode transition in leader-follower arm control~\cite{wu2025robocopilot,evorl2026}. 
As shown in Fig.~\ref{fig:intervention}, the operator first aligns the wrist orientation and hand gesture with the end-effector orientation and dexterous-hand posture predicted by the policy before taking over. Human intervention is then triggered via a keyboard command. Once triggered, the system records the current tracker pose $\mathbf{T}^{v}_{t_0} \in SE(3)$ and the current robot end-effector pose $\mathbf{T}^{\mathrm{ee}}_{t_0}$ as the teleoperation reference poses. At each subsequent timestep $t$, the relative tracker motion 
$\Delta \mathbf{T}^{v}_{t} = \left(\mathbf{T}^{v}_{t_0}\right)^{-1}\mathbf{T}^{v}_{t}$ is computed and applied to the robot end-effector reference pose, resulting in $\mathbf{T}^{\mathrm{ee}}_{t} = \mathbf{T}^{\mathrm{ee}}_{t_0}\Delta \mathbf{T}^{v}_{t}$.
Meanwhile, the policy-predicted hand action is replaced by the retargeted glove action for dexterous-hand control, with smoothing applied during the transition to reduce abrupt action changes at the switching moment. 

\textbf{Staged DAgger-based Data Collection.}
In long-horizon manipulation tasks with sparse rewards, the critic must infer the contribution of each action from delayed terminal outcomes, making it difficult to assign reliable credit to early states over long temporal horizons. To address this question, we introduce staged DAgger, where rollouts are initialized from both the initial state and selected intermediate stages for DAgger-style data collection. Specifically, after an autonomous rollout fails, we restore the environment to the same stage and intervene before the failure occurs, collecting paired failed and corrected trajectories. Such paired trajectories provide contrasting supervision for critic learning around failure-prone states. More importantly, trajectories initialized from later stages provide more direct return supervision for intermediate states, improving their value estimates and serving as reliable intermediate anchors. These anchors shorten the effective credit-assignment horizon for earlier states, enabling value information to be propagated more reliably to earlier task stages without requiring explicit stage labels or task decomposition.

\subsection{Bridging Demonstration-Deployment Gap}
\label{sec:async inference}
\begin{figure}
    \centering
    \includegraphics[width=\linewidth]{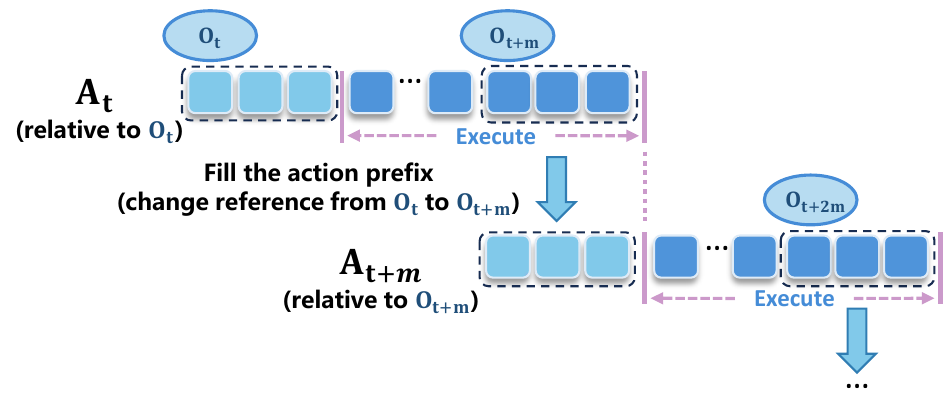}
    \vspace{-2ex}
    \caption{\textbf{Future-state-referenced relative action padding.}
    After asynchronous inference is triggered, the remaining actions in $A_t$ are transformed from the reference frame of $o_t$ to the current observation $o_{t+m}$ and used as the relative-action prefix of the next chunk $A_{t+m}$.}
    \label{fig:async_inference}
    \vspace{-1ex}
\end{figure}

Improving training–inference consistency substantially enhances policy performance, as rollouts more faithfully reflect the behavioral distribution learned during training~\cite{yu2026kai0}. We further argue that this improved fidelity reduces the distribution shift between rollout and demonstration data, allowing the critic to learn a value function induced by a more consistent underlying policy and thereby enabling more accurate policy evaluation.
Specifically, we extend training-time RTC~\cite{black2025training} to a future-state-referenced relative action padding scheme, enabling inference whose execution pattern better matches the action streams seen during training. Let $A_t = [a_t, a_{t+1}, \cdots, a_{t+H-1}]$ denote an action chunk of horizon $H$, and let $n$ denote the maximum inference delay. 
During training, we enable prefix conditioning with probability $p_{m}^{\mathrm{pre}}$, where we uniformly sample a prefix length $i \sim \mathcal{U}\{0,\ldots,n\}$ and condition the policy on the first $i$ actions of $A_t$. This encourages the policy to remain flexible under varying amounts of historical action context, allowing it to handle variable-length prefixes caused by inference delay or intervention recovery. During deployment, once the current action chunk $A_t$ reaches the maximum-delay horizon, asynchronous inference is initiated with the current observation $o_{t+m}$. As shown in Fig.~\ref{fig:async_inference},
The remaining actions of $A_t$, originally relative to $o_t$, are transformed into the reference frame of $o_{t+m}$ and used as the relative-action prefix of the next chunk $A_{t+m}$. While these remaining actions are executed, the policy predicts $A_{t+m}$ asynchronously. Once $A_t$ is completed, the prefix is removed from $A_{t+m}$, and the remaining actions are executed, yielding an inference pattern consistent with the action-prefix conditioning used during training.

\subsection{Policy Improvement from Deployment Experience}
\label{sec:policy improvement}
We next introduce how to perform policy evaluation on an offline dataset composed of autonomous rollouts and human interventions, and how to construct an optimality indicator from advantage signals for policy improvement.

\textbf{Policy Evaluation.}
We adopt the progress-based reward used in $\pi^{*}_{0.6}$~\cite{intelligence2025pi06} as our reward design:
\begin{equation}
r_t =
\begin{cases}
0, & t = T \ \text{and success}, \\
-C_{\mathrm{fail}}, & t = T \ \text{and failure}, \\
-1, & \text{otherwise},
\end{cases}
\label{eq:reward_function}
\end{equation}
where $C_{\mathrm{fail}}$ is set to the maximum episode length of each task.
Its sparsity introduces a credit assignment problem. 
To mitigate this issue, staged DAgger provides empirical returns at different levels of task progress, which alleviates the difficulty of directly using Monte Carlo returns for supervision in long-horizon tasks. Meanwhile, it also introduces heterogeneous data by collecting trajectories from diverse initial states, behavior policies, and outcome qualities, which naturally induces multimodal value distributions. To address both properties, we employ a distributional value network to model the full value distribution~\cite{bellemare2017distributional,intelligence2025pi06}. 
The value is obtained by computing the expectation of the output distribution. 
Using $R_t(\tau)=\sum_{t'=t}^{T} r_{t'}$ to denote the empirical return of a trajectory $\tau$ from time step $t$ until termination, considering the continuous and multimodal nature of value data, we model each empirical return as a Gaussian distribution and discretize it into a soft target over $B$ value bins. 
Specifically, for each empirical return $R_t(\tau)$, we construct a Gaussian soft label over the discretized bins as
\begin{equation}
    q_b(\tau,t)
    =
    \frac{
    \exp\left(
    -\frac{\left(v_b - R_t(\tau)\right)^2}{2\sigma^2}
    \right)
    }{
    \sum_{j=1}^{B}
    \exp\left(
    -\frac{\left(v_j - R_t(\tau)\right)^2}{2\sigma^2}
    \right)
    },
    \quad b=1,\dots,B,
\end{equation}
where $\sigma$ controls the smoothness of the soft target distribution. The resulting vector
\begin{equation}
    \mathbf{q}(\tau,t) = \left[q_1(\tau,t), q_2(\tau,t), \dots, q_B(\tau,t)\right]
\end{equation}
serves as the distributional supervision target for value learning. The distributional critic $p_{\phi}(V \mid o_t)$ is then trained to match this Gaussian soft target by minimizing the cross-entropy loss:
\begin{equation}
    \mathcal{L}_{critic}(\phi)
    =
    \mathbb{E}_{(\tau,t)\sim\mathcal{D}}
    \left[
    H\left(\mathbf{q}(\tau,t), p_{\phi}(V \mid o_t)\right)
    \right],
\end{equation}
where $H(\cdot,\cdot)$ denotes the cross-entropy between the Gaussian soft label and the predicted value distribution.

\textbf{Optimality-Conditioned Policy Improvement.}
Based on the formulation and conclusions in Sec~\ref{sec:Preliminaries}, policy improvement is guaranteed as long as the constructed optimality function $f$ is non-negative and monotonically increasing. Unlike RECAP, which relies on binary optimality labels
$f_\mathrm{bin}\left(A^{\pi_{\mathrm{ref}}}(o_t,a_t)\right)
=
\mathbf{1}\left[
A^{\pi_{\mathrm{ref}}}(o_t,a_t) > q_{\mathrm{low}}
\right]$~\cite{intelligence2025pi06}, 
we construct $f$ as a continuous optimality function,
\begin{equation}
f\!\left(A^{\pi_{\mathrm{ref}}}(o_t,a_t)\right)
=
sig\left(
\frac{
\alpha\left(A^{\pi_{\mathrm{ref}}}(o_t,a_t)-q_{\mathrm{low}}\right)
}{
q_{\mathrm{high}}-q_{\mathrm{low}}
}
\right),
\end{equation}
where $sig(\cdot)$ denotes the sigmoid function, $\alpha$ is a temperature coefficient controlling the sharpness of the mapping, and $q_{\mathrm{low}}$ and $q_{\mathrm{high}}$ denote dataset-level quantiles of the advantage estimates, which mitigates the effect of the absolute advantage scale. Unlike exponential mappings that concentrate the learning signal on a few high-advantage samples, the sigmoid function provides a more robust mapping due to its smooth and bounded nature. Continuous optimality conditioning preserves the relative ordering of action quality instead of collapsing diverse behaviors into binary labels, allowing the policy to exploit fine-grained behavioral differences. 
We condition the diffusion policy on the continuous optimality value and train the actor with the following objective:
\begin{equation}
\begin{aligned}
\mathcal{L}_{\mathrm{actor}}
&=
\mathbb{E}_{\mathcal{D},\eta}
\left[
\left\|
\boldsymbol{\epsilon}
-
\boldsymbol{\epsilon}_{\theta}
\left(
\tilde{\mathbf{a}}_{t:t+h}, o_t, I_t, \eta
\right)
\right\|_2^2
\right], \\
&\text{where }
I_t
=
f\left(A^{\pi_{\mathrm{ref}}}(o_t,a_t)\right).
\end{aligned}
\label{eq:actor_loss}
\end{equation}
where $(o_t,\mathbf{a}_{t:t+h})\sim\mathcal{D}$, 
$\eta\sim\mathcal{U}\{1,\ldots,T\}$ denotes a sampled diffusion timestep, and 
$\boldsymbol{\epsilon}\sim\mathcal{N}(\mathbf{0},\mathbf{I})$ is Gaussian noise. 
The noised action $\tilde{\mathbf{a}}_{t:t+h}$ is obtained by applying the DDPM~\cite{ho2020denoising} forward noising process at timestep $\eta$. During training, we randomly mask the optimality indicator $I_t$ with probability $p_{m}$, replacing it with a null condition. 
This design supports both direct policy sampling conditioned on $I_t=1$ and CFG through the optimality indicator during inference.

\begin{figure}
    \centering
    \includegraphics[width=\linewidth]{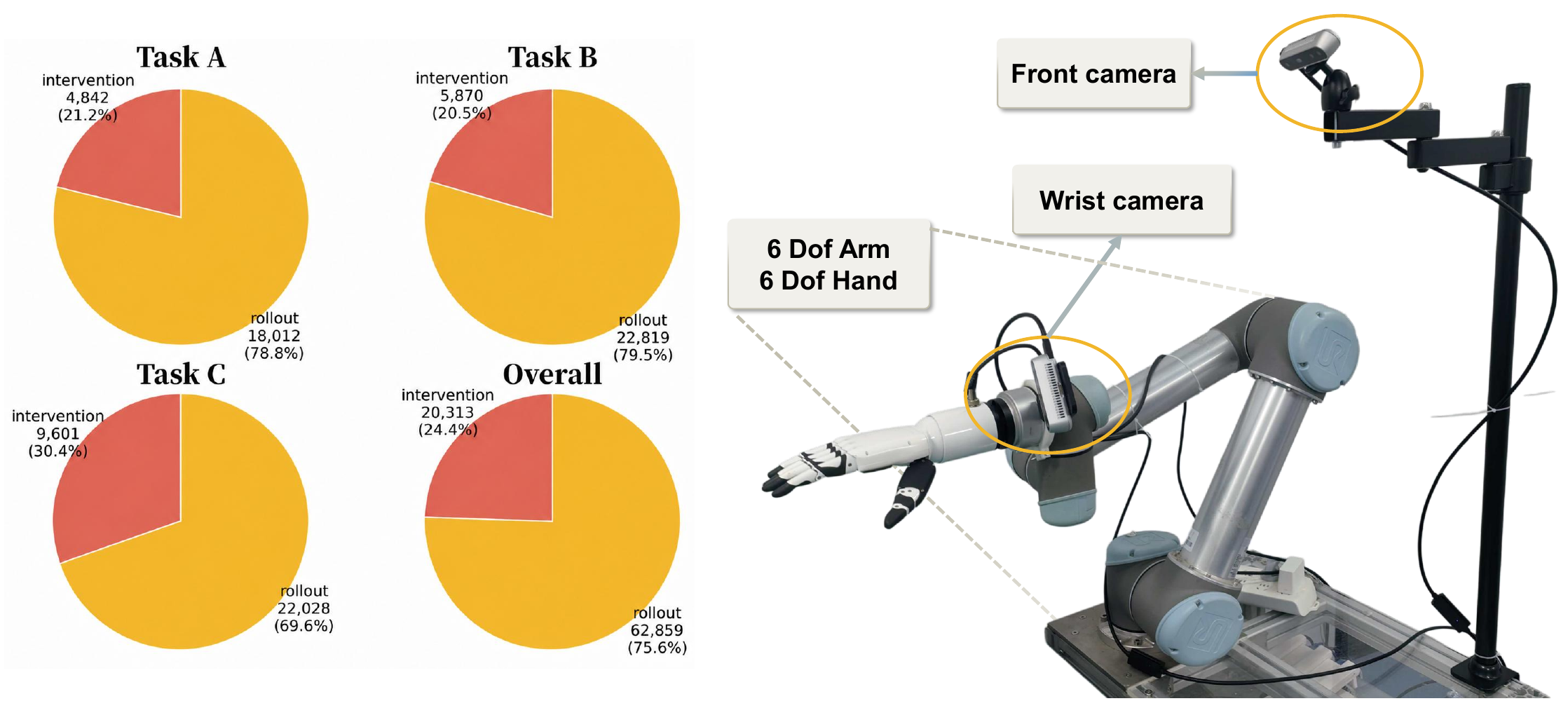}
    \vspace{-3ex}
    \caption{\textbf{Post-training data composition and robot setup}}
    \label{fig:robot_data}
    \vspace{-3ex}
\end{figure}
\begin{figure*}[t]
    \centering
    \includegraphics[width=\textwidth]{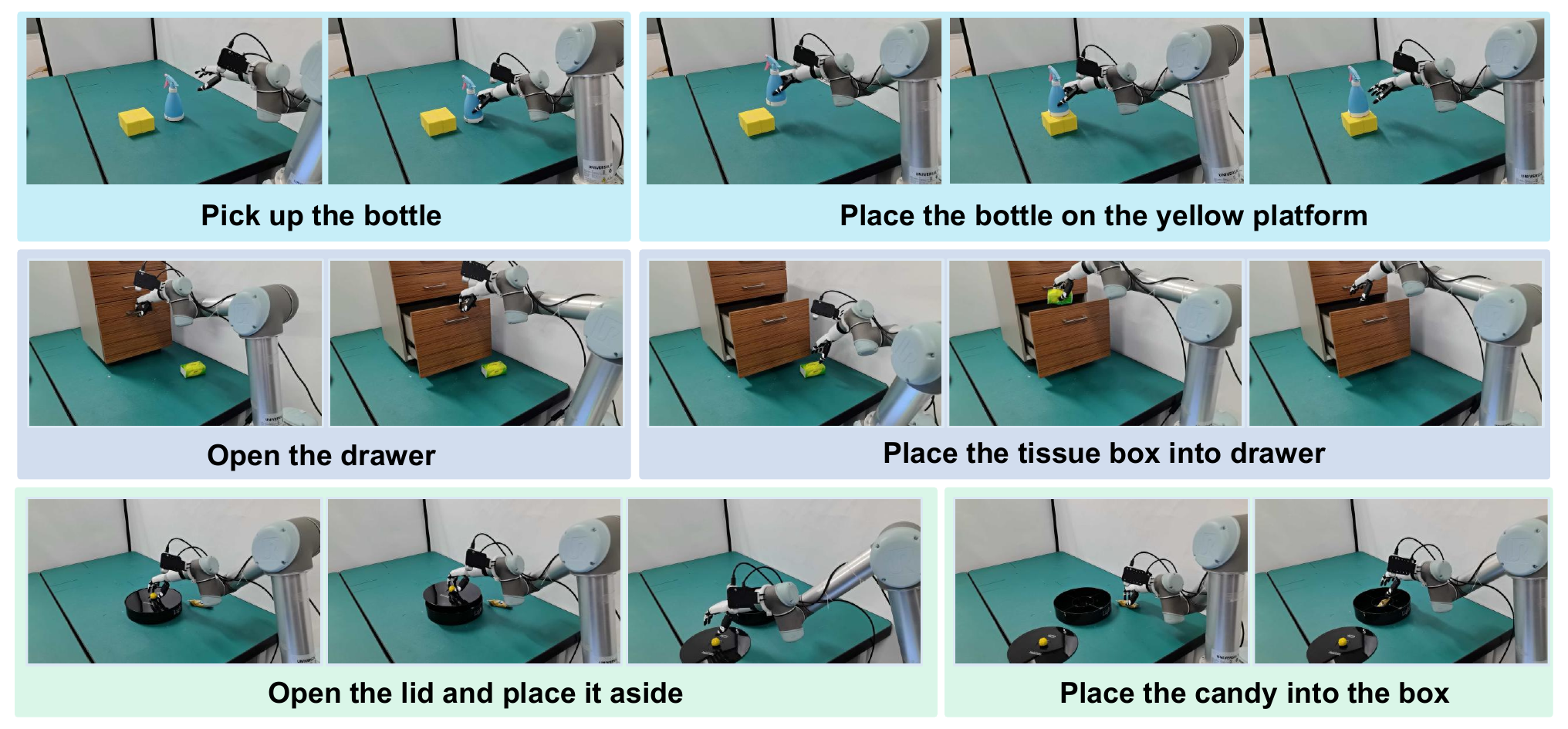}
    \vskip-1.5ex
    \caption{\textbf{Illustrations of the dexterous manipulation tasks.} 
    \textbf{Task A (top)} requires the robot to grasp a tapered bottle and finely adjust its pose for placement. 
    \textbf{Task B (middle)} requires the robot pull the drawer open and place the tissue box inside. 
    \textbf{Task C (bottom)} requires the robot to manipulate the spherical handle to open the lid, and place the candy into the box.}
    \label{fig:task_description}
    \vskip-3ex
\end{figure*}

\section{Experiments}
We design our experiments to answer the following questions:
(1) How does reliable policy evaluation and policy improvement  affect the ability to exploit mixed suboptimal and failure experience?
(2) How does the demonstration--deployment gap affect policy evaluation?
(3) How does staged DAgger affect long-horizon credit assignment?

\textbf{Experimental Setup.} 
As shown in Fig.~\ref{fig:robot_data}, our experimental platform is built upon an arm-hand system consisting of a 6-DoF UR5 robot arm and a 6-DoF Inspire hand with both front-view and wrist-view cameras for visual observation. Our main benchmark includes three real-world robotic manipulation tasks: \textbf{Task A} (Pick-and-Place Bottle), \textbf{Task B} (Open Drawer and Tissue Placement), and \textbf{Task C} (Lid Opening and Candy Placement), as shown in Fig.~\ref{fig:task_description}, with each policy evaluated over $50$ independent trials per task. 
For the average progress score, each trajectory receives $0.5$ for intermediate-stage completion and an additional $0.5$ for full-task completion, and the scores are averaged over $50$ trials. 
For our method, we evaluate the policy after a single iteration of post-training data collection and policy update. 
As shown in Fig.~\ref{fig:robot_data}, we report the proportions of intervention and rollout frames in the collected data. Specifically, we collect $61/37/59$ trajectories for Tasks A/B/C, including $17/15/27$ intervention trajectories and $13/11/15$ failed trajectories, totaling $55.4$ minutes of interaction data.

\textbf{Comparisons and Baselines.} 
We compare against two baselines, RECAP~\cite{intelligence2025pi06} and HG-DAgger~\cite{kelly2019hg}. We collect 41/30/43 demonstration trajectories for Tasks A/B/C, respectively, and use them to train a BC reference policy with the diffusion policy objective~\cite{chi2025diffusion}. The resulting BC policy is then used to warm-start our method and all baselines. All post-training methods are trained on the same post-training dataset, except HG-DAgger, which uses only successful trajectories. 
RECAP differs from our method only by replacing continuous optimality with binary labels and adopting the binary embedding from CFGRL~\cite{frans2025diffusion}, whereas HG-DAgger adopts the same diffusion policy architecture as BC and otherwise follows the same training and inference settings as our method. 

\begin{figure*}[!t]
    \centering
    \includegraphics[width=1.0\textwidth]{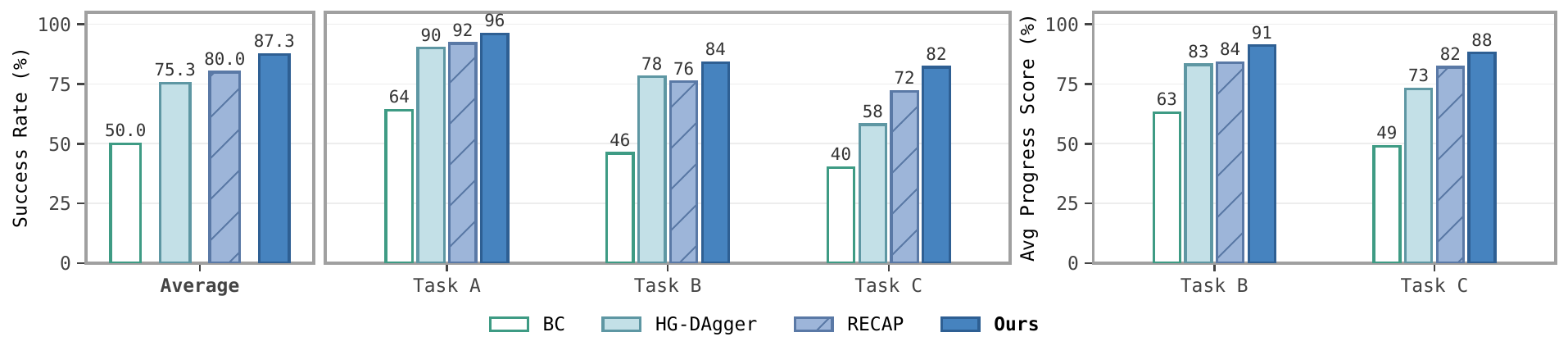}
    \captionsetup{skip=1pt}
    \caption{\textbf{Main results.}
    Comparison with the BC~\cite{chi2025diffusion} and post-training baselines on three dexterous manipulation tasks.
    All post-training methods improve over BC, while our method consistently outperforms HG-DAgger~\cite{kelly2019hg} and RECAP~\cite{intelligence2025pi06}.}
    \label{fig:reult}
    \vspace{-3ex}
\end{figure*}

\begin{figure}
    \centering
    \includegraphics[width=\linewidth]{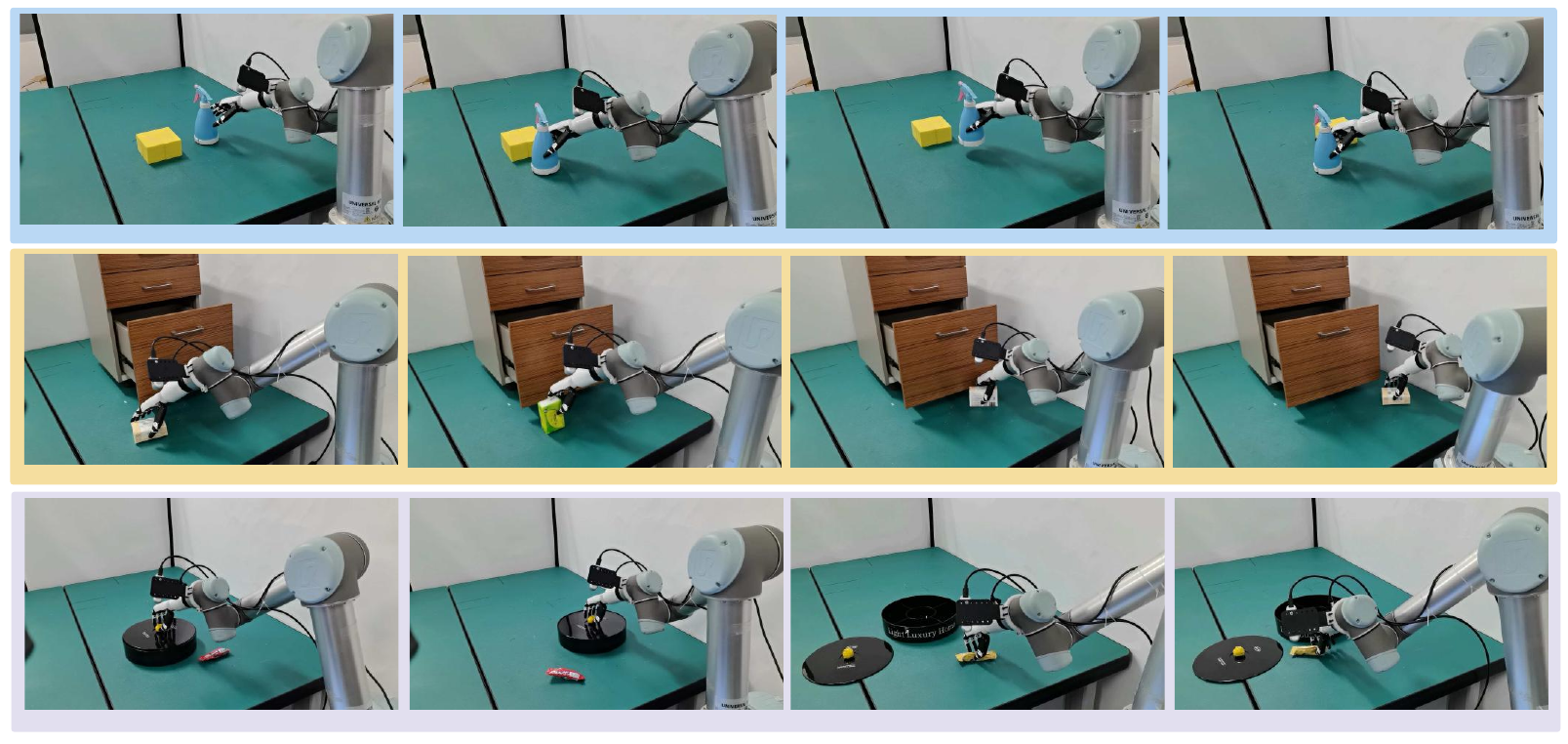}
    \vspace{-3ex}
    \caption{\textbf{Qualitative examples of robustness to positional variations across three tasks.}}
    \label{fig:robustness}
    \vspace{-4ex}
\end{figure}

\begin{figure*}
    \centering
    \begin{minipage}[t]{0.49\textwidth}
        \centering
        \includegraphics[width=\linewidth]{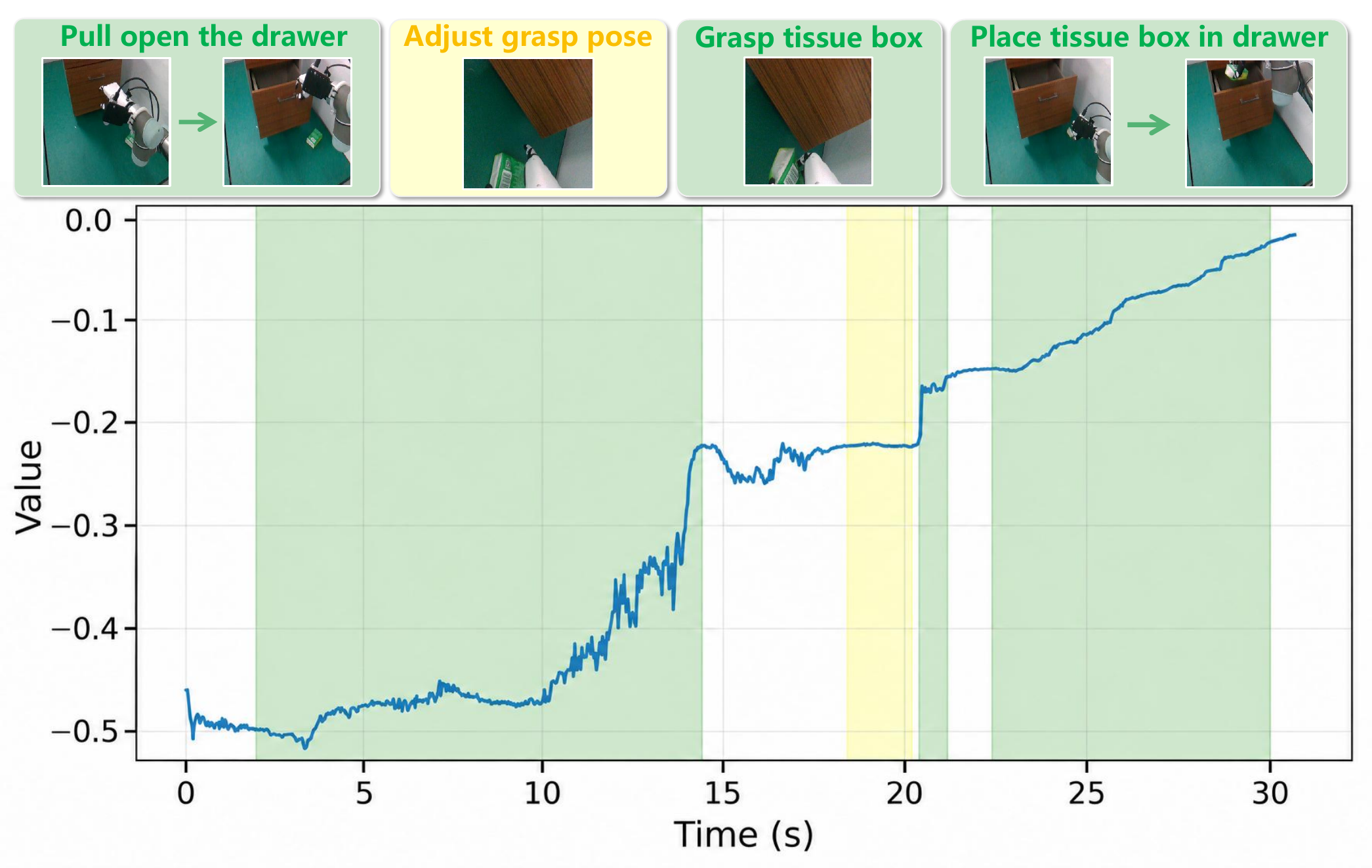}
    \end{minipage}
    \hfill
    \begin{minipage}[t]{0.49\textwidth}
        \centering
        \includegraphics[width=\linewidth]{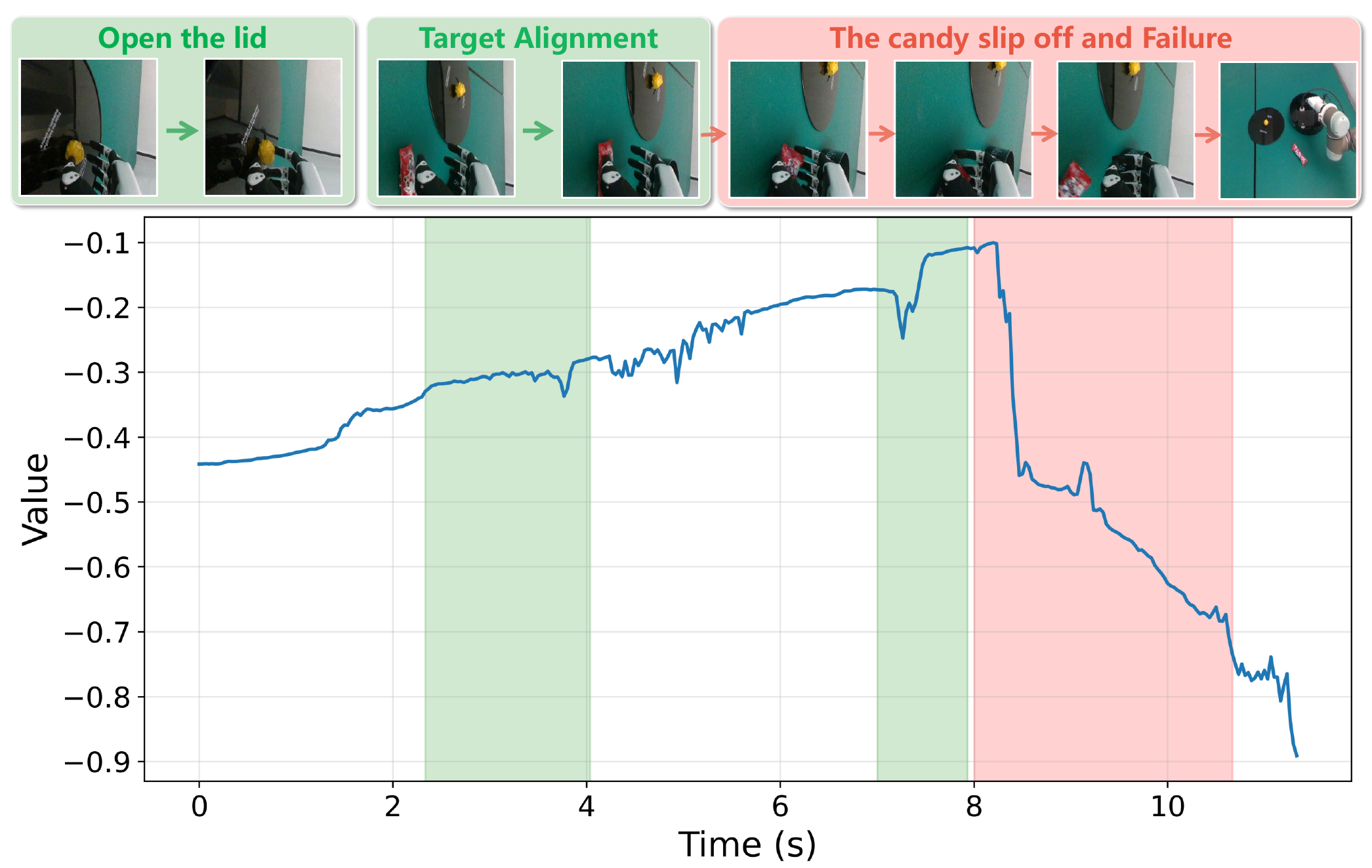}
    \end{minipage}
    \vspace{-1ex}
    \caption{\textbf{Visualization of the value functions.}
    We visualize the value function outputs for one successful and one failed episode, together with keyframes at the corresponding moments.
    The green regions highlight successful progress toward task completion,
    the yellow regions indicate hesitation, and the red regions highlight failure-related segments.}
    \label{fig:value_visualization_1}
    \vspace{-3ex}
\end{figure*}

\textbf{Main Results.} 
As shown in Fig.~\ref{fig:reult}, Our method achieves the largest average improvements over BC in both success rate and task progress, with gains of 37.3\% and 33.5\%, respectively. 
Compared with HG-DAgger, RECAP and our method additionally incorporate failed rollouts into post-training. 
The improved performance with additional failure rollouts shows that reliable policy evaluation can effectively exploit mixed suboptimal and failure experience. By assigning credit at the state level, it can correctly identify and exploit both progressive and failure-inducing behaviors, even within failed trajectories. As shown in Fig.~\ref{fig:value_visualization_1}, we visualize the value curves of two episodes. Benefiting from reduced distribution shift in rollout data and staged DAgger, the value function captures task progress, hesitation, and failure moments, providing reliable credit assignment. 
Meanwhile, compared with the binary labels used in RECAP, our continuous optimality function yields consistent improvements. This demonstrates that continuous optimality conditioning captures action quality more finely than binary labels, thereby enabling more effective policy improvement from mixed-quality suboptimal experience. As show in Fig.~\ref{fig:robustness}, we also observe that our method exhibits stronger robustness to positional variations. This improvement mainly stems from learning both failure cases induced by shifted grasping targets and the corresponding corrective interventions. With reliable credit assignment, suboptimal and failure experience, together with corrective data, jointly improve the coverage of the successful-execution distribution.

\textbf{How does the demonstration-deployment gap affect policy evaluation?} 
We first ablate the design choices that ensure consistency between training and deployment.
As shown in Tab.~\ref{tab:rtc_ablation}, combining training-time prefix conditioning with inference-time prefix padding achieves the highest success rate and the lowest RMS jerk, while maintaining an execution time close to that of demonstrations.
These results indicate that consistency between training and deployment is important for faithfully reproducing the behavior distribution learned from demonstrations.
This motivates us to further investigate whether the resulting demonstration--deployment gap propagates to policy evaluation and ultimately affects downstream policy improvement.
We conduct an ablation study on Task B to examine this effect.
Given the same BC reference policy trained with training-time RTC, we collect post-training rollouts using either synchronous or asynchronous inference. We then perform one iteration of our method on each rollout dataset and evaluate both resulting policies under the same asynchronous inference setting, isolating the effect of rollout collection.
As shown in Tab.~\ref{tab:ablation_async}, the rollout distribution shift decreases the success rate from $84\%$ to $68\%$ and the average progress score from $91\%$ to $78\%$, highlighting its detrimental effect on downstream policy improvement.
To quantify its impact on policy evaluation, we evaluate the critics on 35 unseen Task B trajectories (25 successes and 10 failures) using success/failure accuracy~\cite{chen2025sarm}, explained variance, Spearman's $\rho$, and RMSE.
As shown in Tab.~\ref{tab:critic_ablation}, the critic trained on asynchronous rollouts consistently performs better across all metrics. This suggests that bridging the demonstration--deployment gap allows the critic to learn a value function induced by a more consistent underlying policy, rather than by a heterogeneous mixture of mismatched behaviors, thereby enabling more reliable credit assignment.

\begin{table}[t]
    \centering
    \caption{\textbf{Ablation on training--inference consistency.}}
    \label{tab:rtc_ablation}
    \vspace{-1ex}
    
    \resizebox{\linewidth}{!}{
    \begin{tabular}{lccc}
        \toprule
        \textbf{Method} 
        & \textbf{Success Rate}
        & \textbf{Execution Time (s)$\downarrow$}
        & \textbf{RMS $Jerk_{TCP}^{pos}$ (m/s$^3$)$\downarrow$} \\
        \midrule
        Demonstration
        & -- 
        & 32.1
        & 1.860 \\
        
        Synchronous Inference 
        & 6/20 
        & 46.7 
        & -- \\

        Async w/o Prefix Padding
        & 2/20 
        & 32.3 
        & 8.378 \\

        Async w/o Prefix Training
        & 5/20 
        & 32.8 
        & 7.152 \\

        \textbf{Ours}
        & \textbf{46\%} 
        & \textbf{33.6} 
        & \textbf{4.270} \\
        \bottomrule
    \end{tabular}}
    \vspace{-2ex}
\end{table}

\begin{table}[!t]
    \centering
    \caption{\textbf{Ablation on distribution shift of rollout data.}}
    \label{tab:ablation_async}
    \vskip-1ex
    \setlength{\tabcolsep}{5pt}
    \begin{tabular}{lcc}
        \toprule
        Rollout Data & Success Rate & Avg. Progress Score \\
        \midrule
        Data w/ Sync  & 68\% & 78\% \\
        Data w/ Async & \textbf{84\%} & \textbf{91\%} \\
        \bottomrule
    \end{tabular}
\end{table}

\begin{table}[!t]
    \centering
    \caption{\textbf{Impact of rollout distribution shift on critic evaluation.}}
    \label{tab:critic_ablation}
    \vskip-1ex
    \setlength{\tabcolsep}{4pt}
    \begin{tabular}{lcccc}
        \toprule
        Rollout Data
        & S/F Acc. $\uparrow$
        & Explained Var. $\uparrow$
        & Spearman $\rho$ $\uparrow$
        & RMSE $\downarrow$ \\
        \midrule
        Data w/ Sync
        & 82.85\%
        & 0.6444
        & 0.8785
        & 0.1291 \\
        Data w/ Async
        & \textbf{91.40\%}
        & \textbf{0.7949}
        & \textbf{0.9251}
        & \textbf{0.0863} \\
        \bottomrule
    \end{tabular}
\end{table}

\begin{table}[!t]
    \centering
    \caption{\textbf{Ablation on staged DAgger.}}
    \label{tab:ablation_staged}
    \vskip-1ex
    \setlength{\tabcolsep}{6pt}
    \begin{tabular}{lcc}
        \toprule
        Method & Success Rate & Avg. Progress Score \\
        \midrule
        w/ DAgger & 74\% & 79\% \\
        w/ Staged DAgger & \textbf{82\%} & \textbf{88\%} \\
        \bottomrule
    \end{tabular}
    \vspace{-4ex}
\end{table}

\begin{figure}[t]
    \centering
    \includegraphics[width=0.98\linewidth]{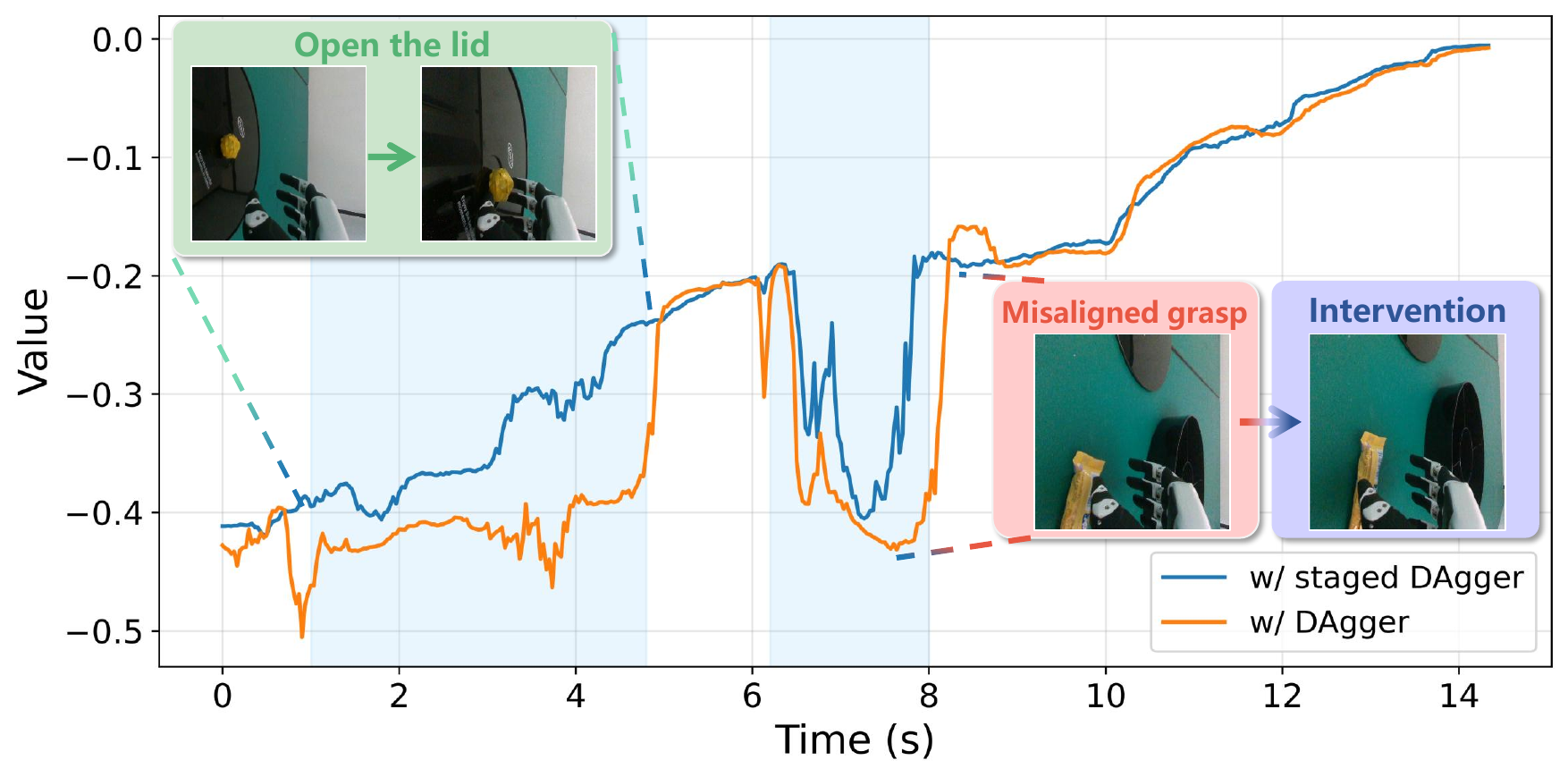}

    \includegraphics[width=0.98\linewidth]{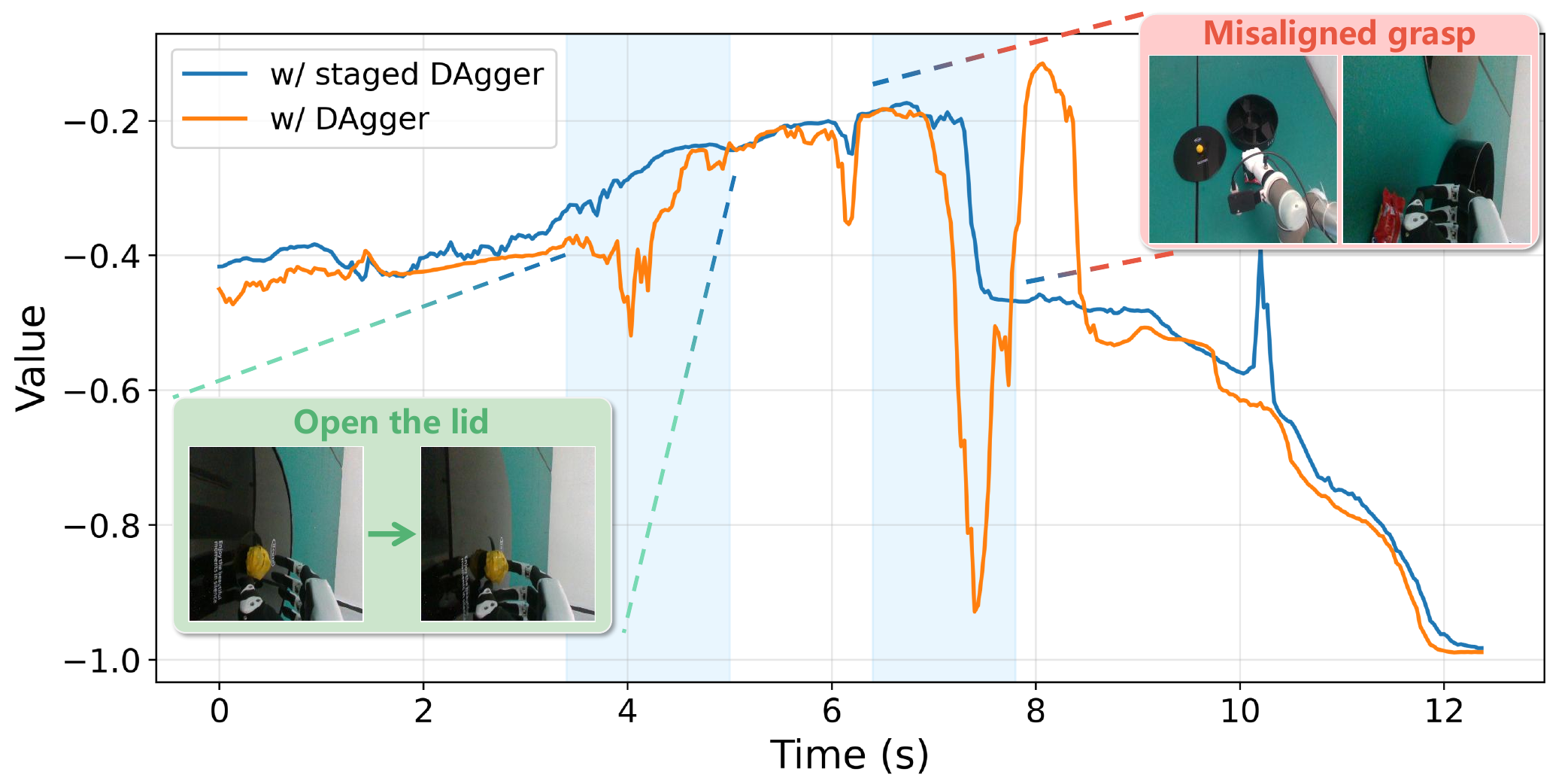}
    \vspace{-1ex}

    \caption{\textbf{Visualization of the value estimates with staged DAgger and with DAgger.}}
    \label{fig:ablation_staged_value}
    \vspace{-2ex}
\end{figure}

\begin{figure}[t]
    \centering
    \includegraphics[width=0.98\linewidth]{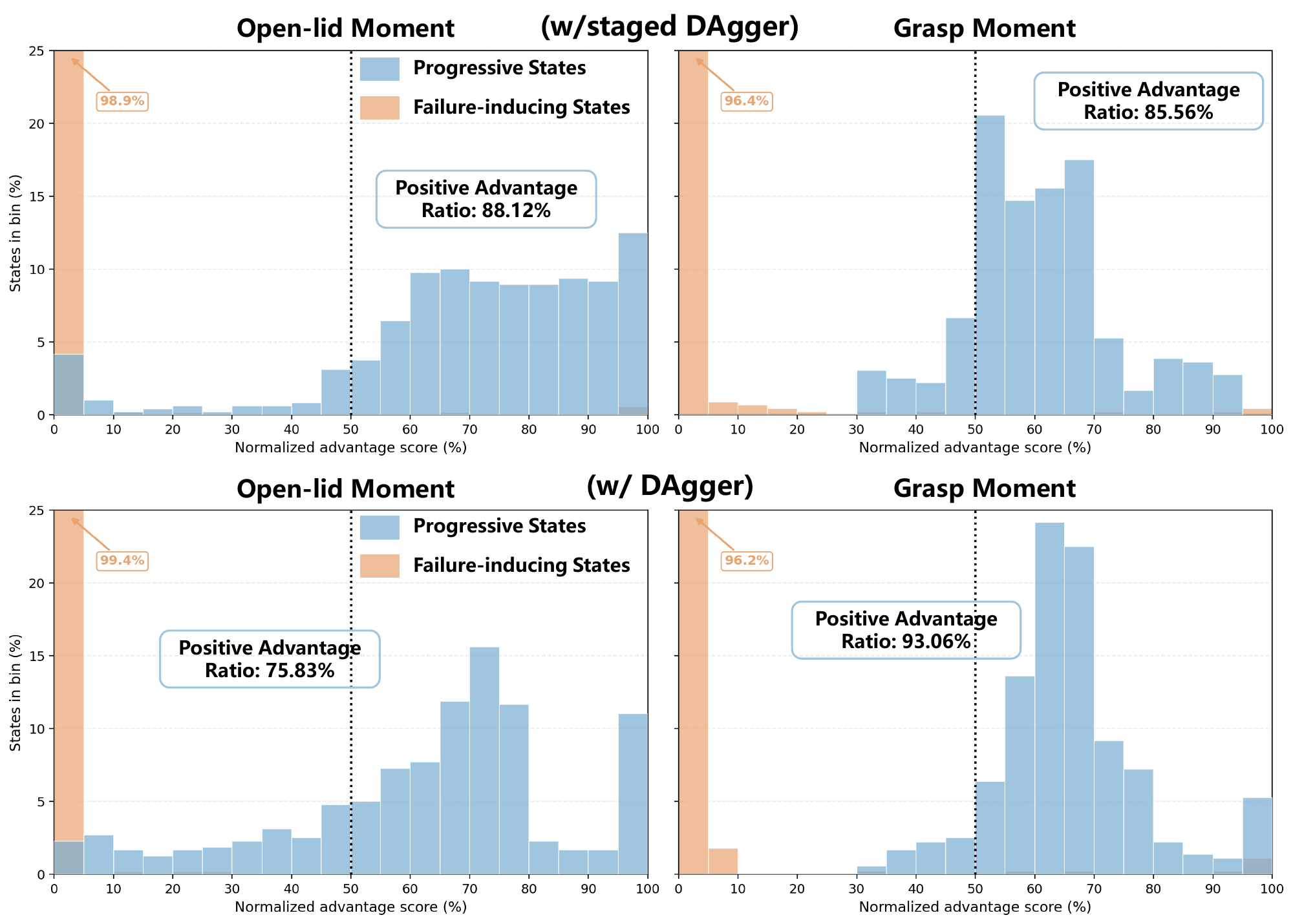}
    \vspace{-1ex}
    \caption{\textbf{Advantage distributions of progressive and failure-inducing states at two critical moments.}}
    \label{fig:advantage}
    \vspace{-4ex}
\end{figure}

\textbf{How does staged DAgger affect long-horizon credit assignment?}
To answer this question, we first evaluate whether staged DAgger facilitates value-function learning and thereby improves policy performance by conducting an ablation study on Task C.
Starting from the same reference policy, standard DAgger collects rollouts only from the initial state, whereas staged DAgger additionally collects trajectories initialized from later-stage states. For a fair comparison, we approximately match the total amount of post-training data by replacing a small portion of the initial-stage trajectories with these later-stage trajectories.
As shown in Tab.~\ref{tab:ablation_staged}, staged DAgger yields an $8\%$ improvement in success rate and a $9\%$ improvement in average progress score, confirming its benefit to downstream policy performance. 
To further investigate its effect on credit assignment, we compare critics trained with standard DAgger and staged DAgger, which extends standard DAgger by incorporating later-stage trajectories. As shown in Fig.~\ref{fig:ablation_staged_value}, both critics capture the late-stage failure and recovery actions reasonably well. At the earlier open-lid stage, DAgger underestimates correct progress, whereas staged DAgger yields progressively higher values. 
This suggests that later-stage data reduces pessimistic credit assignment from downstream failures to earlier correct behaviors. We further quantify this effect in Fig.~\ref{fig:advantage}. Specifically, at two representative moments (open-lid and grasp), we analyze states within approximately $2s$ windows of progressive behaviors and failure-inducing behaviors, and report the distributions of their assigned normalized advantages. Both settings assign mostly low advantages to failure-inducing states, while staged DAgger increases the positive-advantage ratio of progressive states at the earlier open-lid moment by $12.19$ percentage points, correctly attributing positive credit to more early-stage progressive states.
Therefore, we hypothesize that the later-stage trajectories collected by staged DAgger reduce the variance in estimating intermediate-stage values, providing more reliable intermediate value anchors for earlier states and thereby shortening their effective credit-assignment horizon.
\section{CONCLUSION}
We presented \methodname{}, a post-training framework for dexterous policy improvement from real-world experience.
Experiments on three real-world dexterous manipulation tasks show that \methodname{} achieves consistent improvements over the reference policy and competitive post-training baselines. 
Meanwhile, its effective policy evaluation provides quality annotations for autonomous experience, thereby unlocking the potential of incorporating these real-world post-training data into large-scale pre-training.  
Due to hardware and computational constraints, our evaluation is limited to single-arm manipulation tasks. 
Future work should explore policy improvement in bimanual manipulation settings, as well as its effectiveness on minute-level long-horizon tasks.

{
\small
\bibliographystyle{ieeetr}
\bibliography{main}
}

\end{document}